# IDENTIFYING INTERACTION SITES IN "RECALCITRANT" PROTEINS: PREDICTED PROTEIN AND RNA BINDING SITES IN REV PROTEINS OF HIV-1 AND EIAV AGREE WITH EXPERIMENTAL DATA

MICHAEL TERRIBILINI[1,3†], JAE-HYUNG LEE[1,3], CHANGHUI YAN[1,2,4], ROBERT L. JERNIGAN[1,3,5], SUSAN CARPENTER[6], VASANT HONAVAR[1,2,5], DRENA DOBBS[1,4,5]

*[1]Bioinformatics and Computational Biology Graduate Program and L.H. Baker Center for Bioinformatics and Biological Statistics, [2]Department of Computer Science, [3]Department of Biochemistry, Biophysics and Molecular Biology, [4]Department of Genetics, Development and Cell Biology, [5]Artificial Intelligence Research Laboratory, and Center for Computational Intelligence, Learning and Discovery, Iowa State University, Ames, IA, 50010, USA*

*[6]Department of Veterinary Microbiology and Pathology, Washington State University, Pullman, WA, 99164, USA*

Protein-protein and protein nucleic acid interactions are vitally important for a wide range of biological processes, including regulation of gene expression, protein synthesis, and replication and assembly of many viruses. We have developed machine learning approaches for predicting which amino acids of a protein participate in its interactions with other proteins and/or nucleic acids, using only the protein sequence as input. In this paper, we describe an application of classifiers trained on datasets of well-characterized protein-protein and protein-RNA complexes for which experimental structures are available. We apply these classifiers to the problem of predicting protein and RNA binding sites in the sequence of a clinically important protein for which the structure is not known: the regulatory protein Rev, essential for the replication of HIV-1 and other lentiviruses. We compare our predictions with published biochemical, genetic and partial structural information for HIV-1 and EIAV Rev and with our own published experimental mapping of RNA binding sites in EIAV Rev. The predicted and experimentally determined binding sites are in very good agreement. The ability to predict reliably the residues of a protein that directly contribute to specific binding events - without the requirement for structural information regarding either the protein or complexes in which it participates - can potentially generate new disease intervention strategies.

† Corresponding author

## 1. Introduction

The human AIDS virus, Human immunodeficiency virus Type 1 (HIV-1), is closely related to a number of lentiviruses that cause persistent, insidious infections in other primates and domestic animals. Recent advances in molecular virology have resulted in novel antiviral therapies that inhibit specific proteins required for the replication of lentiviruses and other important retroviruses. Rev is a multifunctional regulatory protein that plays an essential role in the production of infectious virus (1, 2) and, as such, is an attractive target for new antiviral therapies. To date, however, no Rev-targeted drugs for AIDS therapy are available.

Rev is known to participate in protein-protein interactions with several cellular proteins as well as in RNA-protein interactions with lentiviral RNAs (3, 4). It is required for the transition to the late stage of viral replication and facilitates export of incompletely spliced viral RNAs from the nucleus to the cytoplasm. After its import into the nucleus, HIV-1 Rev binds a structure in the viral pre-mRNA called the Rev-responsive element (RRE) (5, 6), multimerizes (6, 7), then utilizes the CRM1 nuclear export pathway to redirect movement of incompletely spliced viral RNA out of the nucleus (8). As shown in Figure 1, functional domains within HIV-1 Rev are known to mediate interactions with viral RNA and with host cell proteins that are required for nuclear localization, RNA binding, multimerization, and nuclear export (3).

Efforts to develop inhibitors of Rev activity have been hampered by a lack of information regarding Rev protein structure. A major stumbling block for structural analysis is the tendency of Rev to aggregate at concentrations needed for crystallization or solution NMR studies (9). The only high resolution information available is for short peptide fragments of HIV-1 Rev. In an NMR solution structure of a 23 amino acid fragment of Rev bound to a 34 nucleotide RRE RNA fragment, the Rev peptide adopts an $\alpha$-helical conformation and is bound in the major groove of the RNA (10). Structures of other critical functional domains of Rev (e.g., nuclear localization, multimerization, export) have not been reported. Furthermore, it has not been possible to apply homology modeling approaches to gain insight into Rev structure because Rev has no detectable sequence similarity to any protein of known structure. Indeed, despite their apparently conserved functions, Rev protein sequences are highly variable between species, with < 10% sequence identity between HIV-1 and one of the most divergent Rev proteins, equine infectious anemia virus, (EIAV) Rev (11).

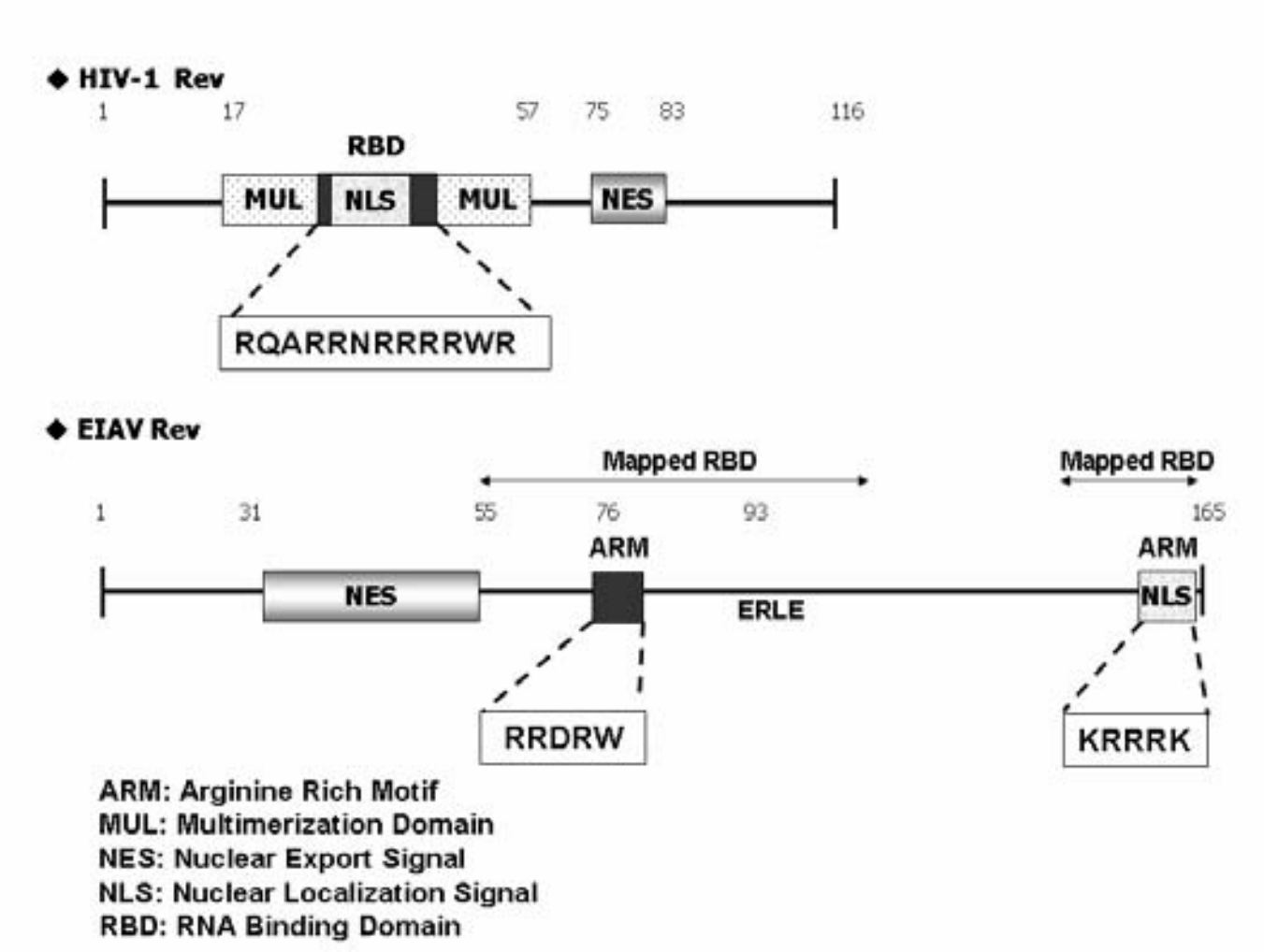

**Figure 1.** Functional domains of HIV-1 and EIAV Rev proteins. The linear organization of functional domains within the two Rev proteins differs significantly, but both have been shown to contain specific sequences involved in Rev interactions with proteins (MUL, NLS, NES) or RNA (RBD, ARMs).

When protein structures cannot be solved using experimental approaches, computational analyses can provide valuable insight into protein structure-function relationships and aid in identification of key functional residues that may offer tractable targets for therapeutic intervention in disease (12). Here we describe the identification of critical residues that mediate protein-protein and protein-RNA interactions in Rev, using machine learning approaches that rely on the primary amino acid sequence of Rev, but do not require any information regarding its structure or the sequence or structure of its interaction partners. Our predictions are in good agreement with previously published biochemical, biophysical and genetic data for HIV-1 and EIAV Rev as well as with our recent experimental mapping of RNA binding sites in EIAV Rev (13). Taken together, these results demonstrate the utility of sequence-based approaches for identifying putative binding sites of proteins with potential therapeutic value that are, at present, recalcitrant to experimental structure determination.

## 2. Datasets, Materials and Methods

### 2.1. *Datasets*

***Protein-protein binding site dataset (PBS).*** We extracted individual proteins from a set of 70 protein–protein heterocomplexes used in the study of Chakrabarti and Janin (14). After removal of redundant proteins and molecules with fewer than 10 residues, we obtained a dataset of 77 individual proteins with sequence identity <30%. The dataset contains a total of 12,719 amino acids, of which 2340 (18.4%) are interface residues (positive examples).

***RNA-protein binding dataset (RBS).*** A dataset of protein-RNA interactions was extracted from structures of known protein-RNA complexes in the Protein Data Bank (PDB) (15). Proteins with >30% sequence identity or structures with resolution worse than 3.5Å were removed using PISCES (16). This resulted in a set of 109 non-redundant protein chains containing a total of 25,118 amino acids. Amino acids in the protein-RNA interface were identified using ENTANGLE (17). Using default parameters, 3518 (14%) of the amino acids in the dataset are defined as interface residues (positive examples).

### 2.2. *Protein-protein interface residue prediction*

We have previously developed a two-stage classifier for predicting interface residues in protein-protein complexes (18). In the first stage, a Support Vector Machine (SVM), trained on the PBS dataset, is used to classify each residue as interface or non-interface. Input to the SVM is a window of nine amino acid identities. Because interface residues tend to be clustered in primary sequence, a second stage was introduced to take advantage of this to improve predictions. In the second stage, a Bayesian Network classifier is trained based on the predictions of the target residue and its neighbors from the first stage SVM. Let $C \in \{0,1\}$ denote the actual class label of a residue; $X \in \{0,1\}$ be the prediction of the SVM classifier; Y denote the number of predicted interface residues within 4 amino acids of the target residue. For each residue, the likelihood that it is an interface residue given the SVM predictions for itself and its neighbors is calculated and compared to a chosen threshold $\theta$ as formula 1.

$$\frac{P(C=1 \mid X,Y)}{P(C=0 \mid X,Y)} > \theta \qquad (1)$$

The residue is predicted to be an interface residue if the likelihood is larger than $\theta$ and non-interface otherwise. The conditional probability table P(C|X,Y) is derived from training datasets. To determine $\theta$, the classifier was applied to the training set and different values of $\theta$ ranging from 0.01 to 1 were tested, in increments of 0.01. The value of $\theta$ for which the classifier yields the highest correlation coefficient was used to make predictions on the Rev proteins.

### 2.3. ***Protein-RNA interface residue prediction***

We have previously developed a Naïve Bayes (NB) classifier for predicting which amino acids in a given protein are likely to be found in protein-RNA interfaces (19), using the NB classifier from the Weka package (20). The input is a window of 25 contiguous amino acid identities. The output is an instance where + indicates that the target residue is an interface residue and – indicates a non-interface residue. A training example is an ordered pair (x, c) where $x = (x_{-n}, x_{-n+1}, ..x_{T-1}, .x_T, x_{T+1}, ...x_{n-1}, x_n)$ and c is the corresponding class label (interface or non-interface). A training data set D is a collection of labeled training examples.

Let $X = (X_{-n}, ...X_T, ...X_n)$ denote the random variable corresponding to the input to the classifier and C denote the binary random variable corresponding to the output of the classifier. The Naïve Bayes classifier assigns input x the class label + (interface) if:

$$\frac{P(C=+)\prod_{i=-n}^{i=n} P(X_i = x_i \mid C=+)}{P(C=-)\prod_{i=-n}^{i=n} P(X_i = x_i \mid C=-)} \geq \theta$$

and the class label – (non interface) otherwise. $\theta$ was set to the value that optimized the correlation coefficient (21) on the *training set* in each leave-one-out cross validation experiment.

### 2.4. *Experimental mapping of RNA binding sites*

Details of our experimental mapping of RNA binding sites are provided in Lee et al., (13). Briefly, Maltose Binding Protein-EIAV Rev (MBP-ERev) constructs containing deletions or point mutations in the EAIV Rev coding region were cloned in pHMTc, based on the pMal-c2x expression vector, which enhances solubility of Rev fusion proteins. MBP-ERev fusion proteins were expressed in *E. coli*, purified prior to use in RNA binding experiments. UV cross linking experiments were used to quantitate the effects of mutations on Rev RNA binding activity (13).

## 3. Results

### 3.1. *Binding site predictions on datasets of known protein-protein and protein-RNA complexes*

In previous work, we have developed classifiers for predicting interface residues in protein-protein, protein-DNA and protein RNA complexes (18, 19, 22), typically using a combination of sequence and structure-derived information as input. In choosing classifiers for the task of predicting protein-protein and protein-RNA interface residues in Rev proteins, we compared several types of classifiers for predicting each type of interface residue (data not shown). Table 1 shows an example of the classification performance values obtained for protein binding site prediction using the PBS dataset, which contains 77 proteins used in our previous study (18) and for RNA binding site prediction using the RBS dataset, which contains 109 RNA-binding proteins (19).

**Table 1.** Classification performance in predicting protein-protein and RNA-protein binding site residues, using leave-one-out experiments

| Classification Performance Measure | Protein Interface Residues (2-stage classifier) | RNA Interface Residues (NB classifier) |
|---|---|---|
| **Accuracy** | 72% | 85% |
| **Specificity** | 58% | 51% |
| **Sensitivity** | 39% | 38% |
| **Correlation coefficient** | 0.30 | 0.35 |

These results were obtained using a modified 2-stage classifier developed in this work to predict protein interface residues (see Methods) and a Naive Bayes classifier published previously (19) to predict RNA interface residues. The results of the latter study are reproduced here for comparison.

### 3.2. *Predicted binding sites in wildtype HIV-1 and EIAV Rev proteins*

Using classifiers trained on the datasets described above, we predicted protein-protein and protein-RNA interface residues in Rev proteins from HIV-1 and EIAV. As shown in Figure 2A, the 2-stage protein classifier predicted a total of 56 protein-protein interface residues (indicated by "**p**") within the 116 amino acid HIV-1 Rev sequence. These are primarily located in 5 clusters consisting of 6-15 amino acids. The Naive Bayes classifier predicted a total of 26 RNA-protein interface residues (indicated by "**r**"), located in a single large cluster near the N-terminus of the protein. The predicted RNA binding site sequence is PPNPEGTRQARRNRRRRWRERQRQIHSIG, corresponding to amino acids 28-56. Ile26 and Ile29 are the only two residues within this sequence that are predicted to be non-interface residues.

The prediction results for EIAV Rev, using the same classifiers, are shown in Figure 2B. A total of 79 protein-protein interface residues were predicted in the 165 amino acid protein. In EIAV Rev, most of these predicted protein-binding residues are also located in 5 clusters that are somewhat larger (8-24 amino acids) than those predicted in HIV-1. There are two predicted clusters of RNA-protein interface residues, one consisting of 15 contiguous amino acids, located in the central region and a second consisting of 19 contiguous residues at the C-terminus of the protein. The predicted RNA binding site sequences are RHLGPGPTQHTPSRR, (aa 63-77) and QSSPRVLRPGDSKRRRKHL (aa 147-165. The only other predicted interface residues are 5 scattered amino acids in the region of aa 113-133.

### 3.3. *Comparison of predicted Rev binding sites with experimental data*

Functional domains in HIV-1 Rev have been extensively interrogated through the analysis of sequence variants and mutants generated both *in vivo* and *in vitro* (4). These experimental results are summarized in Figure 1 and mapped onto amino acid sequence of HIV-1 Rev for comparison with our predicted RNA and protein interface residues in Figure 2A. Notably, the single cluster of RNA interface residues predicted by the Naive Bayes classifier closely matches the experimentally mapped RNA binding domain (RBD), which in HIV-1 also includes an Arginine Rich Motif (ARM) that also functions as a nuclear localization signal (NLS). Three predicted clusters of protein interface residues also characterized protein binding sites: one cluster (aa 22-32) maps to Rev multimerization domain, and two clusters are located within a large C-terminal domain (aa 87-116) that has been shown to play multiple roles in nuclear export, dimerization and transactivation activities of HIV-1 Rev (23). One of these clusters (aa75-93) also overlaps with the modular nuclear export signal (NES), which is interchangeable between various lentiviruses, including HIV-1 and EIAV (24).

Although the functional domains in EIAV Rev have been studied in less detail than those in HIV-1 Rev, previous biochemical and genetic studies had localized the NLS and NES domains and implicated two motifs in the central region in RNA binding, RRDRW and ERLE (Figure 1) (13, 25-28). In predictions generated *before* we initiated our experimental mapping of EIAV RNA-binding domains, the Naïve Bayes classifier identified one potential RNA-binding region overlapping the RRDRW motif and another overlapping a KRRRK motif within the mapped C-terminal NLS domain, but did not predict any interface residues near the ERLE motif. Our recent direct mapping of the RNA binding domain of EIAV Rev by UV cross linking showed that two separate regions of Rev are necessary for RNA binding: a central region encompassing aa 75-127 and a region comprising the 20 C-terminal residues of EIAV Rev (13). These experiments also demonstrated critical roles for both the central RRDRW motif and the KRRRK motif within the NLS in RNA-binding (13). Interestingly, however, the ERLE motif was not required for RNA-binding, in agreement with our predictions. Thus, our biochemical RNA-binding site mapping studies for EIAV Rev have provided direct experimental validation of the RNA interface residue predictions of the Naive Bayes classifier.

Of the five clusters of predicted protein binding residues in EIAV Rev, two overlap with known or putative protein interaction domains (the NES and the NLS, respectively), one is located in the non-essential "hypervariable" region (13), one is located near the N-terminus of the protein, and one overlaps within the central RNA binding domain (Figure 3B). There is no available biochemical data regarding the possibility that the central region of EIAV Rev binds both RNA and protein, but it is interesting that the classifier predicted binding of the NLS region to both protein and RNA. The same residues could directly interact with both the nuclear import machinery and RNA because these interactions occur at different times and in different cellular compartments. Also, by analogy with HIV-1 Rev, it is likely that some of the protein interactions that occur when EIAV Rev multimerizes after binding RNA involve additional residues located near the RNA binding region that initiates the specific interaction between Rev and the RRE in unspliced EIAV RNA.

**A.**

```
        1.........11........21........31........41........51........
SEQ     MAGRSGDSDEELIRTVRLIKLLYQSNPPPNPEGTRQARRNRRRRWRERQRQIHSISERIL
PRO     ....pppppppp.........ppppppppppp...........................p
RNA     ...r.........................r.rrrrrrrrrrrrrrrrrrrr.rr.r.r..

        61........71........81........91........101.......111...
SEQ     GTYLGRSAEPVPLQLPPLERLTLDCNEDCGTSGTQGVGSPQILVESPTVLESGTKE
PRO     .pppppppppp.....p..pppppppppppppp.pp..ppp.........pppppp
RNA     ........................................................
```

**B.**

```
        1.........11........21........31........41........51........
SEQ     MAESKEARDQEMNLKEESKEEKRRNDWWKIDPQGPLESDQWCRVLRQSLPEEKISSQTCI
PRO     .....pppppppp..........ppppppppppppppppppp.....pp...........
RNA     ............................................................

        61........71........81........91........101.......111.......
SEQ     ARRHLGPGPTQHTPSRRDRWIREQILQAEVLQERLEWRIRGVQQVAKELGEVNRGIWREL
PRO     ppppppppppppppp....................................ppppppppp
RNA     ..rrrrrrrrrrrrrrr....................................r......

        121.......131.......141.......151.......161..
SEQ     HFREDQRGDFSAWGDYQQAQERRWGEQSSPRVLRPGDSKRRRKHL
PRO     ppppppppppppppp.............pp...ppp...pppppp
RNA     ............r.............rrrrrrrrrrrrrrrrrrr
```

**Figure 2.** Predicted interface residues in Rev proteins. The protein sequences (SEQ) for **A)** HIV-1 Rev & **B)** EIAV Rev are shown on top line, with binding site residues for protein (PRO) and RNA shown by "p" or "r" on the lines below. Important functional domains boxed in the sequence are: NES, NLS/ARM ,RBD, MULTIMERIZATION, MULTIFUNCTIONAL, ARM, UNKNOWN.

### 3.4. *Comparison of predicted and biochemically mapped RNA binding sites in EIAV mutant Rev proteins*

Site-specific mutagenesis, coupled with functional assays, has identified functional domains of EIAV Rev (13, 25, 26). As mentioned above, an NLS/ARM at the C-terminus was identified at the EIAV Rev C-terminus and our cross-linking analyses of the RRDRW and KRRRK motifs indicated that both are likely to contact RNA. To investigate whether our classifiers are capable of detecting mutations that give rise to differences in RNA binding, we performed predictions on several mutant EIAV Rev sequences. As shown in Figure 3, changes in RNA interface predictions are seen in sequences in which Ala residues are substituted for positively charged residues in the RRDRW and KRRRK motifs (to AADAA and KAAAK). These mutations result in >80% reduction in RNA binding activity (13). The predicted RNA binding sites no longer overlap these motifs. In contrast, predicted protein interface residues remain unchanged, consistent with the experimental results.

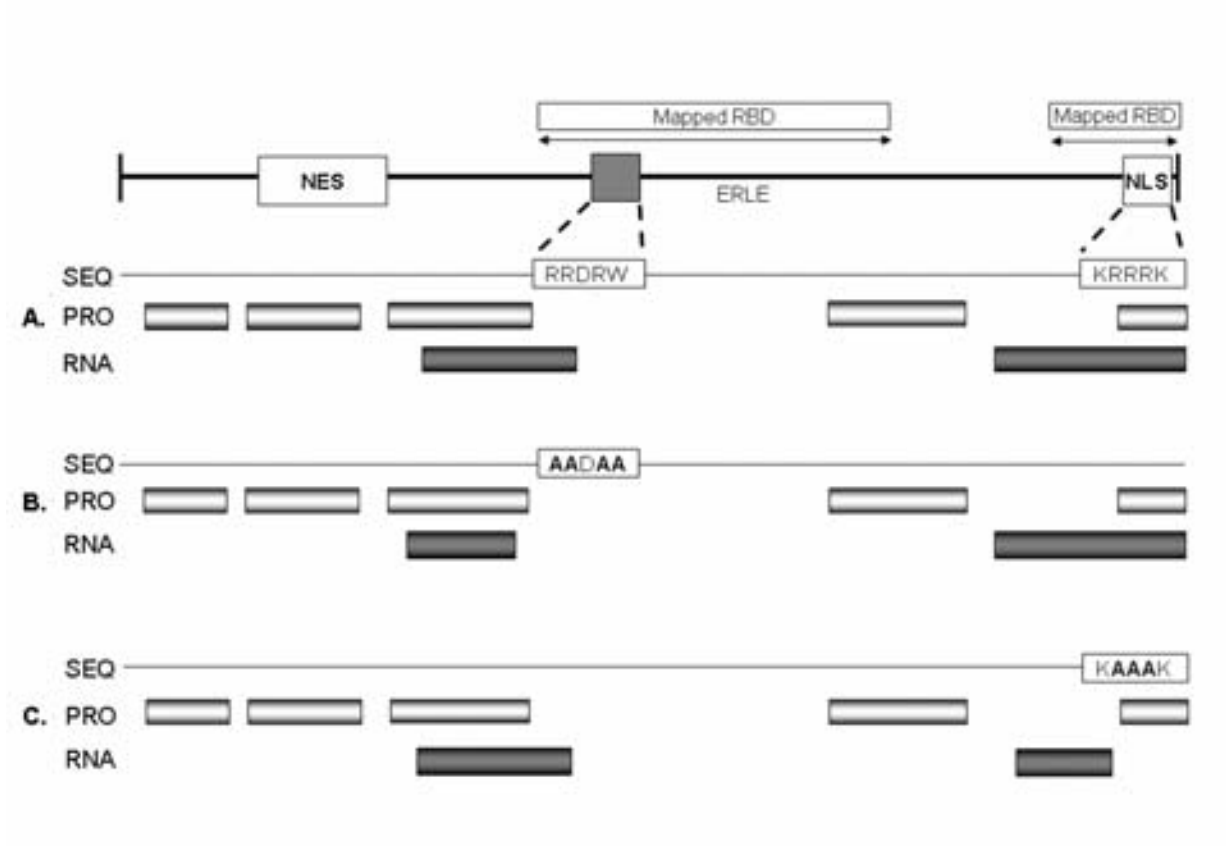

**Figure 3.** RNA binding site predictions differ for "wildtype" and mutant EIAV Rev sequences. Predicted protein (PRO) and RNA binding sites are indicated along the sequence (SEQ). **A**. Wildtype, **B**. & **C**. Mutant EIAV Rev sequences. RNA binding activity is reduced by >80% in both mutants (see text for details).

## 4. Summary and Discussion

Many effective antiviral drugs are directed at blocking the interaction between regulatory proteins and their binding partners or small effector ligands. HIV-1 Rev is one of several clinically important proteins that are "experimentally recalcitrant," i.e., for which it has not been possible to obtain high resolution structural information. Identifying critical functional residues in Rev is further complicated by the fact that Rev proteins have no significant sequence similarity to any protein with known structure, and that Rev sequences from different species have very little similarity to one another.

Our comparison of predictions with experimental data on the Rev proteins from HIV-1 and EIAV demonstrates that sequence-based computational methods can identify residues in "recalcitrant" proteins that interact with other proteins or nucleic acids. When structural information *is* available for a protein of interest, enhanced prediction accuracy can be achieved (18, 29). Developing improved methods for predicting binding sites will contribute to our understanding of how proteins recognize their targets in cells and may significantly decrease the time needed to precisely map binding sites in the laboratory. The level of accuracy obtained using the sequence-based methods presented here suggests that they could expedite the design of experiments to explore the function of key regulatory proteins, even when no structural information is available, with obvious implications for developing new therapies for both genetic and infectious diseases.

## Acknowledgments

This Research was supported in part by grants NIH, GM 066387 (VH, DD, & RLJ) and CA97936 (SC), by an ISU Center for Integrated Animal Genomics grant (DD, VH & RLJ), and by USDA Formula Funds (SC & DD). We thank Sijun Liu for technical assistance and Jeffrey Sander for useful comments.